\documentclass[AMA,STIX1COL]{WileyNJD-v2}

\articletype{Article Type}%

\received{26 April 2016}
\revised{6 June 2016}
\accepted{6 June 2016}

\begin{document}

\title{Brittle AI, Causal Confusion, and Bad Mental Models: Challenges and Successes in the XAI Program} 

\author[1]{Jeff Druce*}

\author[2]{James Niehaus}

\author[1]{Vanessa Moody}

\author[3]{David Jensen}

\author[4]{Michael L. Littman}


\address[1]{\orgdiv{Decision Management Division}, \orgname{Charles River Analytics}, \orgaddress{\state{Massachusetts}, \country{USA}}}

\address[2]{\orgdiv{HCIS Division}, \orgname{Charles River Analytics}, \orgaddress{\state{Massachusetts}, \country{USA}}}

\address[3]{\orgdiv{College of Information and Computer Sciences}, \orgname{University of Massachusetts Amherst}, \orgaddress{\state{Massachusetts}, \country{USA}}}

\address[4]{\orgdiv{Computer Science Department}, \orgname{Brown University}, \orgaddress{\state{Rhode Island}, \country{USA}}}

\corres{Jeff Druce. \email{jdruce@cra.com}}


\abstract[Summary]{The advances in artificial intelligence enabled by deep learning architectures are undeniable. In several cases, deep neural network driven models have surpassed human level performance in benchmark autonomy tasks. The underlying policies for these agents, however, are not easily interpretable. In fact, given their underlying deep models, it is impossible to directly understand the mapping from observations to actions for any reasonably complex agent. Producing this supporting technology to "open the black box" of these AI systems, while not sacrificing performance, was the fundamental goal of the DARPA XAI program. In our journey through this program, we have several "big picture" takeaways: 1) Explanations need to be highly tailored to their scenario; 2) many seemingly high performing RL agents are extremely brittle and are not amendable to explanation; 3) causal models allow for rich explanations, but how to present them isn't always straightforward; and 4) human subjects conjure fantastically wrong mental models for AIs, and these models are often hard to break. This paper discusses the origins of these takeaways, provides amplifying information, and suggestions for future work. }



\maketitle


\section{Introduction}\label{sec1}

In 2016 we were in the midst of what many would call an Artificial intelligence (AI) boom. Due to advances in deep neural networks (DNNs) \cite{goodfellow2016deep}, the state-of-the-art approach for a variety tasks (e.g., image classification, denoising, generative modeling, translation tasks,  autonomy) were being advanced at a  staggering rate. However, one reason for this success is due to the vast architectural complexity of the underlying DNNs at the heart of the models that performed these tasks. Although the fundamental mathematical operations used by DNNs are relatively simple, the vast scale of these operations obfuscates the nature of how the systems produces output. For example, consider a linear model that predicts the rent given $x_1$ the apartment area,  and $x_2$, the number of bedrooms as \emph{features} (e.g.,  $ y(x_1,x_2) = 2.1 x_1 + 0.6 x_2 + 500 $). If we have a few instances of the output, it's straightforward to see how how the rent goes up when the area goes up, with a baseline value for lower areas. However, what if we mapped the input to \emph{latent} features via additional linear models, before mapping to the rent (e.g., $z_1(x_1,x_2) = 0.4x_1 + 0.8x_2 - 232, z_2(x_1,x_2) = 1.4x_1 + 2.8x_2 - 500$, and $y'(z_1, z_2) = 0.2 z_1 - 0.4 z_2 - 125 $). Just by looking at a few instances, it would become much harder to see patterns. Now, in a DNN, there may exist thousands or even millions of these latent features, rendering it virtually impossible to understand how the input is mapped to output. To further complicate matters, many networks have skip connections \cite{mao2016image}, recurrent blocks \cite{zaremba2014recurrent} and convolutional operations \cite{lecun1995convolutional}) that further muddy the waters.

In nearly all cases, this lack of understandability in AIs precludes them from use in critical applications. This need for explainability was a primary motivation for the DARPA XAI program \cite{gunning2017explainable}, whose core goal was to enable the usage of these cutting edge models, while not sacrificing (or even improving) their performance. "Enabling the usage" is a complex, interdisciplinary challenge spanning a full spectrum from computer science to experimental psychology. As a result, the program not only sought to  develop technology to explain how advanced AI systems work, but also how to deliver comprehensible explanations for \emph{users} of the systems, not just the \emph{developers} of the systems. The latter portion led to a variety of fascinating research revealing the challenging nature in getting humans to accept and trust AI \cite{lee2004trust}. Finally, to validate that the explanations were indeed helpful, the program also included a series of user studies, which itself posed many challenges--not the least of which is developing reasonable quantitive measures to test against \cite{hoffman2018metrics}. 

In this paper, as participants in the XAI program, we highlight some interesting lessons learned along the way. We begin with a discussion on when explainability is appropriate and reasonable for AI-driven autonomy, and dig into our choice for human-machine-teaming scenario. We then discuss the hard truth we faced that our seemingly high-performing RL agents simply memorized their action choices in response to particular environment configurations, with no real concept of the makeup or physics of their world \cite{wity_gen} Next, we discuss challenges encountered in producing counterfactual explanations, as well as the difficulties in paring down which counterfactuals should be delivered to a user (counterfactuals can informally be thought of as answers to \emph{what if?} queries). Finally we discuss difficulties in designing AI explanation systems for non-AI experts. Namely, the unforeseen hurdles in overcoming user's erroneous mental models of how AIs work.

\section{Scenarios Appropriate for Explanations}\label{sec2}
A tenet of our work is that we wanted to provide explanations for autonomy based AIs used in critical scenarios. Given the current maturity of AI-driven autonomy, and in particular verification and validation (V\&V) of AI-driven autonomy, we opted for a Human Machine Teaming (HMT) scenario where a human would take on a task alongside an agent (human in the loop). Additionally, we wanted the scenario to be reasonably complex, such that a human would struggle to do this ask alone (due to the \emph{divided attention} nature of the task), and that a modern deep RL agent is required for sufficient (human-like) performance. 

After cutting our teeth with providing explanations for autonomous in the Amidar environment, we began the transition to more complex domains. We selected the domain of StarCraft2 as it supports HMT, allows for custom maps enabling as much or as little complexity as needed and is popular in the AI community \cite{alphastarblog} (hence has ample hooks into the environment such as the PySC2 framework \cite{vinyals2017starcraft}). Critical for our scenario, we wanted the human to have the option of using the AI or not, where it would be beneficial to use the agent in some scenarios, and harmful in others--motivating the user to understand the agent and the scenarios in which they will be helpful. Given the AIs would be so complex that real-time explanations would not be an option, we offer an After Action Review (AAR) interactive training system. 
  
Our first attempt at HMT in StarCraft2 was met with limited success. The complexity of our initial scenario was simply too great; it required highly performing AIs employing micro and macro strategies, and hence required explanations of great breadth and depth for sufficient understanding of the AIs.  Our custom map consists of 2 groups of 5 space marines  defending a base against 3 roaches; the human has the option of taking control of both groups, or giving control of one group to an AI. The roaches spawn 3 at a time around the edges of the map with different health levels of 100, 200, 300 (see Figure \ref{UI}). That wave ends when all roaches are killed or all the Marines and the supply depot are destroyed. At the end of each wave, the following points as given:20 points for each living Marine (up to 200), 1 point for each point of supply depot health (up to 100). 

The results of this evaluation gave clear guidance that there must exist a careful balance between the complexity of the environment, the sophistication of the agent, and the nature of \emph{how} explanations are delivered. In the case above it was clear it would have taken a very sophisticated AI to be successful in our environment. However, in our user study, a user was to have access to the explanation system for approximately one hour, and the user was to not have a background in AI. To sufficiently cover the many nuanced behaviors of the AI, in the many configurations games may take, in that short amount of time in a way that is accessible to a non-AI expert is simply intractable. Hence, it becomes critical to simultaneously consider factors such as how complex the AI is, how complex its environment, how much time a user has access to the system, and the background of the user when designing an XAI framework. 

\section{Look How Smart It Is! Challenges in Training Agents Amendable to Explanation}

Another tenet of work is that a causal understanding of AIs affords explanations that are naturally suited for helping users understand, trust, and effectively use them. To that end, we wanted to train agents where we could \emph{intervene} on the environment (that is, directly change one aspect of the environment while holding everything else constant) and see how the agent would respond. In doing this, we could generate \emph{experimental data} allowing us to more easily learn causal models, and obtain the type of explanations we are seeking. 

\begin{figure*}
\centering
    \includegraphics[width=390pt]{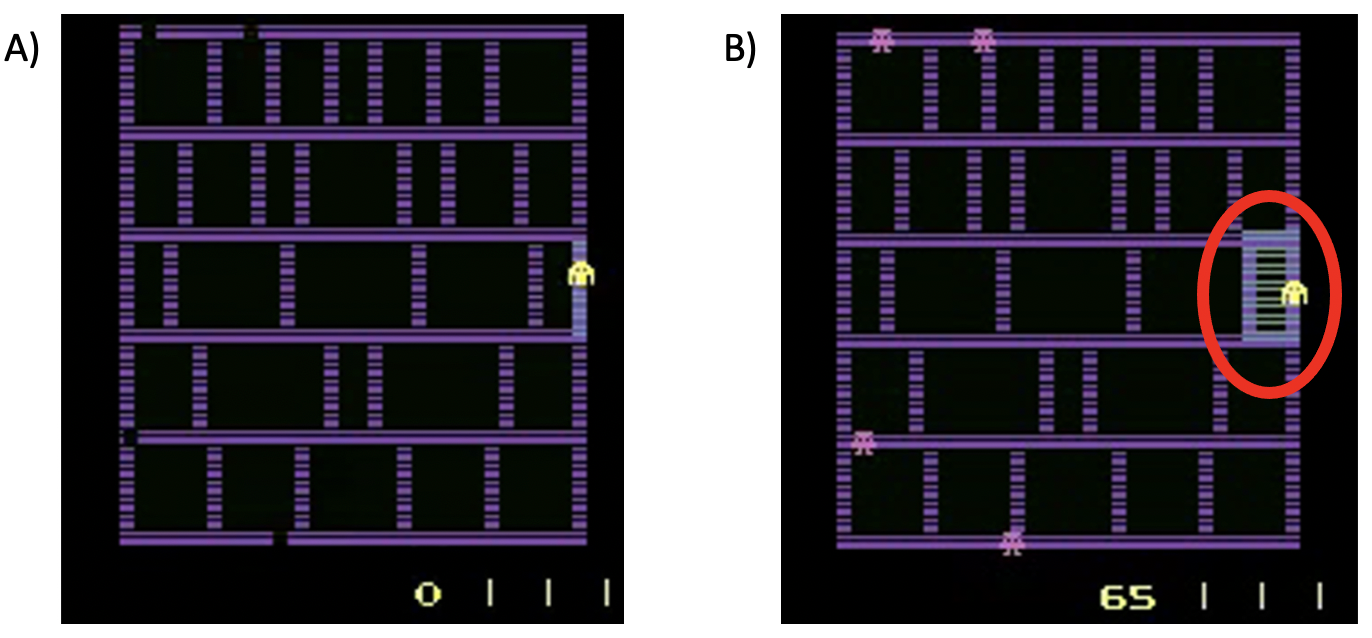}
    \caption{The Brittleness of our Reinforcement Learning trained Amidar Agent. Figure A) shows the unaltered Amidar environment, where the agent performs well (comparable to a human user) appearing to dodge enemies, and efficiently pursue points. Figure B) shows the Amidar environment with a small, seemingly meaningless perturbation: the addition of a "painted" region (naturally occurs during gameplay). The addition results in an Amidar agent performing significantly worse than when in the original environment, even though this level is no more difficult.  }
    \label{agents}
\end{figure*}

Our starting point for exploring causal explanations was to train a deep reinforcement-learning agent to play an Atari video game. Of the 50+ possible games, we chose to focus on Amidar (see Figre \ref{agents}); it was a game that was well handled by existing deep RL algorithms and had sufficiently interesting dynamics that we believed it would support rich explanations. (We spent a week playing 1980s era video games and rating each one on whether it was playable---challenging but not too challenging---and whether it was strategically interesting.) We applied an implementation of DQN to learn a policy for the game. We selected DQN because it was the first deep RL algorithm and it was known to perform well on this game \cite{mnih2013playing}].

After training the agent, to the point where it was consistently finishing all or most of the first level of the game, we set out to understand how it was making the decisions it was making. We could see that it sometimes made some very interesting moves, like moving toward an enemy in the game and then turning onto a “side street” at the last possible moment. It could dodge in and out of enemy traffic, completing the board while it did so.

Given a particular action at a particular time, we wanted to know why it did that. In our way of thinking about explanations, it was important to understand under what natural variations of the present circumstances would it have chosen a different action. We were interested in finding things like (see the figure): “It went west at this point to avoid the enemy. In particular, had the enemy not been there (a counterfactual), the agent would have continued traveling in the direction it was moving.” As another example (see the figure), we wished to be able to say: “The agent went north at this point to visit an unvisited patch of the board. In particular, had that patch already been marked as visited (a counterfactual), the agent would have headed toward a different part of the board.”

Try as we might, we couldn’t get our explanatory mechanism to output sensible explanations. The agents did not appear to have any kind of consistent reaction to the counterfactual alternate situations we posed. After significant amounts of debugging, we began to wonder if the problem was in the agent, not in our explanatory mechanism. 

To test this possibility, we constructed a set of challenges for our trained agent. We ran Amidar with fewer, or even zero, enemies. We ran Amidar with patches of the board already marked as visited (see Figure \ref{agents}. We transported the player to other parts of the board. In all cases, we found that the score obtained by the agent plummeted. The agent had learned reasonable action choices to take in the situations it typically encountered during training and testing, but these decisions were not well generalized to even highly similar states. Simply put, the agent had memorized a set of good choices for some states but had not learned a general rule for assessing possible choices in others. As a result, the actions it was taking were fundamentally unexplainable. Why did it take a certain action in a certain state? Because that’s the action it learned to take in that state. We likened it to an actor in a play. Why did you storm out of the room at that point? Because that’s what it said to do in the script. What would you have done if the situation were different? I have no idea, the script didn’t cover any other situations.

Obviously this kind of behavior is not a necessary property of RL algorithms. We set out to try to understand whether some algorithms were better able to learn generalizing policies than others. We carried out a set of experiments in a variety of well-studied RL environments: CoinRun, CartPole, Amidar, Star Craft II, Lunar Lander, Pong, and Pong with access to the underlying RAM. We compared the value-based method DQN we had begun with to the PPO policy-based algorithm that remains one of the most reliable deep RL algorithms. We also experimented with MuZero \cite{schrittwieser2020mastering}, a deep model-based algorithm that had been demonstrated to play Atari games effectively. We found that all of the algorithms were prone to a kind of overfitting---if we trained and tested them in the same environment, they did fine. If we intervened and tested them in environments that differed, even subtly, from their training environment, they performed poorly.

\begin{figure*}
\centering
    \includegraphics[width=400pt]{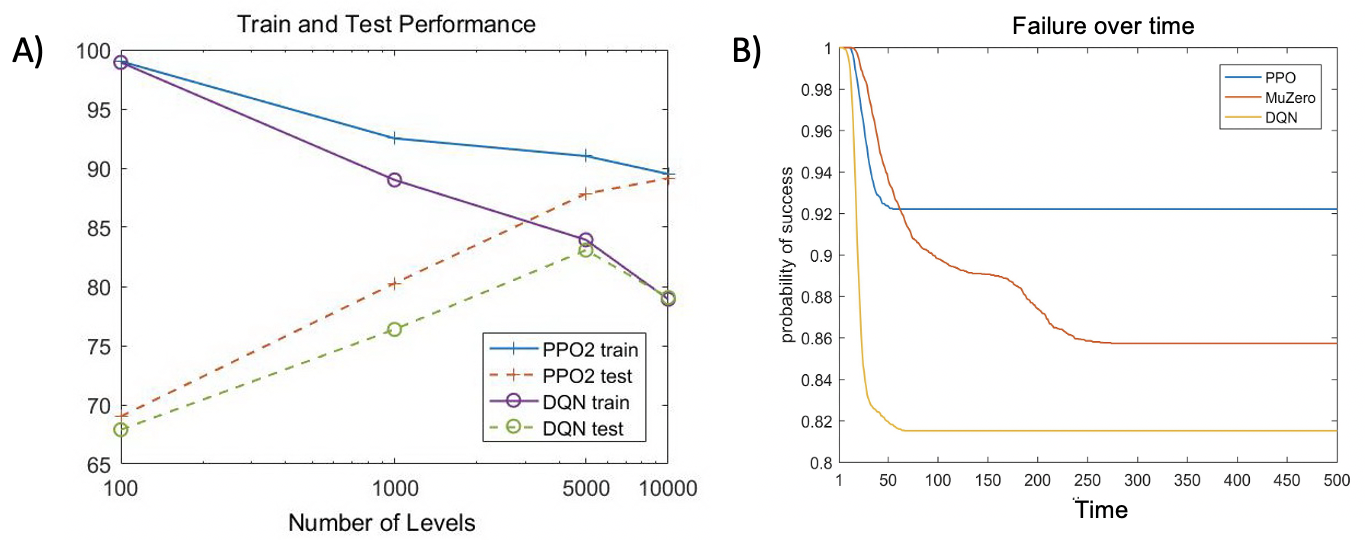}
    \caption{RL agent performance. In Figure A), the $x$ axis represents the number of environment variants encountered by the algorithm during training. After 10,000 variants, PPO2, the policy-based algorithm, generalized to 90\% of novel environment variants, while DQN, the value-based algorithm, generalized to only 80\% of novel environments variants. Figure B) shows the likelihood of failure of 3 different policies. It is clear the DQN is extremely brittle to environmental changes. }
    \label{RL}
\end{figure*}

The CoinRun domain was particularly interesting because it is designed to allow agents to train in a variety of related environments instead of just one \cite{cobbe2019quantifying}. The authors of the domain  had shown that, compared to value-based algorithms, policy-based algorithms tended to require less variability in their training environments to begin to learn robust, consistent behavior that generalized well. We replicated their finding, as shown in the Figure \ref{RL}.

In follow-up experiments, we repeated this experiment in a simpler environment, CartPole, and included MuZero as a third test algorithm. (Our attempts to run MuZero on any more complex environments such as CoinRun, Amidar, or Star Craft II failed due to the computational demands of the algorithm. Deep Mind made good use of Google’s resources in running their experiments and we could not replicate them.) In this test, we wanted to know how well the algorithms could balance a pole when the starting configuration varied far more widely in the test evaluations than in training. As shown in Figure \ref{RL}, PPO again learned the most robust policy---balancing the pole for 500 steps in 92\% of the tests compared to DQN’s 82\%. MuZero’s performance was far more interesting. It was better than the other algorithms at keeping the pole balanced for 50 or fewer steps. But it came out in the middle when tasked with balancing for the full 500 steps. Interestingly, the design of MuZero requires it to make predictions over a fixed predefined window of steps. We had chosen 50 because longer predictions made the algorithm too expensive to run. Thus, the model-based algorithm was more effective (and likely explainable) than the others when it was applicable. But, due to its high costs, it applied to fewer problems than the other algorithms.

\section{Casual Models for Explanations}\label{sec3}
A key challenge in producing trustworthy AI systems is providing counterfactual explanations.  Accurate counterfactual explanations require identifying the inference that a given AI system would have made if the inputs to the system been different.  Identifying such counterfactual conclusions appears simple: merely input the counterfactual conditions and determine the system's inference.  However, the level of description of the counterfactual conditions often do not match those of the system's inputs.  For example, if a computer vision system fails to detect a pedestrian, the desired counterfactual explanation might state that the pedestrian would have been detected if the pedestrian had been wearing brighter clothing.  The level of the desired explanation (clothing choices) does not match the system input (pixels).  Such mismatch has been shown to produce misleading counterfactual conclusions when interventions are made at the level of the system input and then naively used to inform human reasoning about higher-level counterfactuals \cite{atrey2019exploratory}.

\subsection{Wait, who killed my flowers? Plausibility of explanations}

As a result, our approach focused on learning a separate causal model that describes what influences the inferences of an AI system (specifically, an RL policy learned via a deep network).  The inputs to the causal model (in causal parlance, the "treatments") were variables describing the perceptions of the RL agent that were defined at a level that could be used directly in explanations.  The outputs of the causal model ("outcomes") were actions of the RL agent.  In the example above, the causal model might indicate how the height, clothing color, and relative position of a pedestrian influences how accurately a computer vision model identifies them as a pedestrian.

With multiple queries to the learned causal model, an explanation system can infer which counterfactual conditions ("interventions") would change the decision of the RL agent. Almost invariably, however, many different counterfactual conditions can change the agent decision. Relatively large interventions on single conditions can change the agent decision, as can smaller interventions on multiple conditions. One intervention can amplify, or it can hinder, the effect of other interventions.

Thus, an accurate causal model alone is not sufficient for counterfactual explanation.  A key problem remains: How to convey this complex counterfactual landscape to a user with limited time and attention.  One seemingly simple approach is to rank alternative counterfactual conditions by some criterion and provide a user with a ranked list. How that ranking should be performed, however, is far from clear.

Human reasoning typically considers a range of factors when evaluating which of a set of counterfactual conditions are most relevant for explaining a given outcome.  One classic formulation considers how best to explain why the flowers died when both the gardener and the Queen of England failed to water them.  If either had acted to water them, the flowers would have survived, but only the gardener's actions are typically taken as a relevant counterfactual \cite{menzies2001counterfactual}.  However, in the general case, automating such reasoning to provide a ranking over counterfactual conditions remains a research frontier.

\section{User Study}\label{sec4}

\begin{figure*}
    \includegraphics[width=500pt]{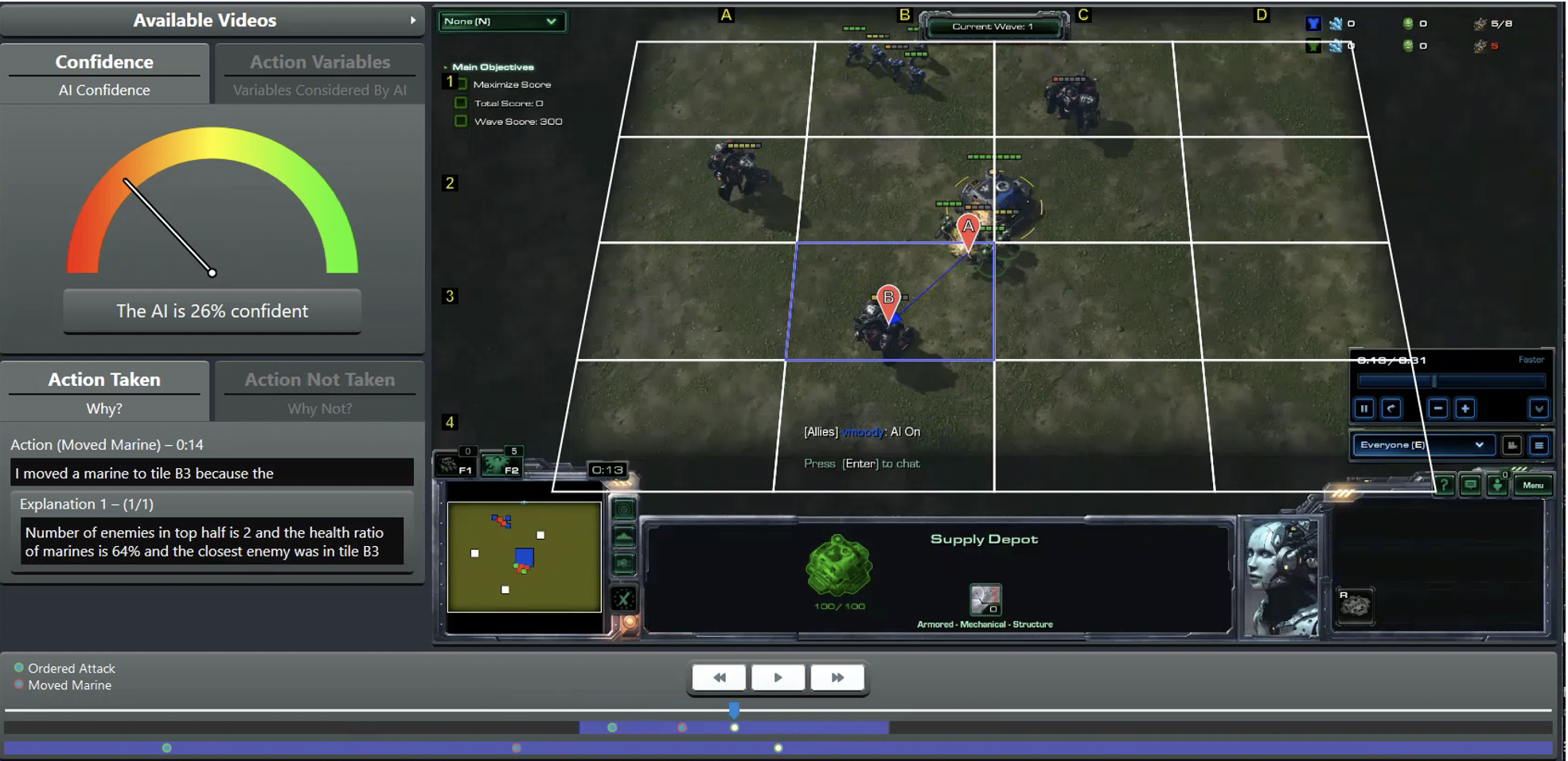}
    \caption{The CAMEL Explanation User Interface (XAI). The top right of the XUI shows a replay of StarCract2 gameplay with overlaid spatial grid. Below the gameplay is a timeline showing all actions taken by the AI; each action can be selected for analysis. In the top left is the "Confidence" plot which shows how well the AI is predicting how well it will perform in this environment. Below this, in the "Action Taken" tile, are counterfactual explanations showing what action the AI took, and the driving environmental factors that led to this (e.g., for the scenario shown here, the agent attacked location B3 due to the number of enemies in the top half, and the marine health value).  The gird allows users to localize these driving factors as they are highlighted on the screen (e.g., Location A shows the current location of the marines, and location B shows where the agent has selected to attack).  }
    \label{UI}
\end{figure*}

A key component of the XAI program was that we were providing explanations for users and not developers of AIs. This created a host of difficulties in delivering explanations concerning extremely technical systems in a non-technical manner, and these difficulties surfaced in our user studies. Our capstone user study paired human participants with Deep RL AI helper agents in a StarCraft2 custom minigame (see Figure \ref{UI}). We recorded how participants performed during the task and their mental models about the AI at baseline (i.e., doing the whole task themselves),with an AI before any explanation (PreX), and with an AI after CAMEL generated explanations (PostX).The participants were able to choose when to use and when not to use the AI. They could also play the game in a way to attempt to maximize the use of the AI’s strengths and minimize the impact of the AI’s weaknesses. Therefore, participants had a strong incentive to understand the AI and use the explanations.The study was carefully balanced to help ensure: the task was difficult enough that the AI could assist, theAI had strengths and weaknesses that impacted the situations in which it was useful, and that training aDRL AI was feasible under the scope of the study. In this study, we found that CAMEL explanations improved participant mental models and performance on the joint AI teaming task, and we did not find evidence  that  mental  models  or  performance  improved  significantly  without  explanation.  Participants were better able to understand when the AI would perform well and when it would perform poorly (as well as the factors that lead to these situations) and enact that understanding to improve performance.

While the full study findings are presented elsewhere, we learned two major lessons during this study that we present here: (1) we needed to overcome impressively wrong mental models and (2) better mental models led to better teaming.

\subsection{Overcoming Impressively Wrong Mental Models}

Participants in our study were recruited to have moderate skill playing StarCraft 2. They had 50-100 hours playing StarCraft 2, are probably gamers in general, and have had significant exposure to commercial game AI (e.g., the AI that controls the enemy units during a single-player campaign). During this experience, they developed significant expectations and biases on what it means to be a game AI and how it should perform. Additionally, because even the StarCraft 2 minigame is complex and dynamic, they are likely to develop simplified folk models based on initial impressions. In real world environments, these rapid and approximate mental models enable people to work together in teams almost instantaneously. 
However, DRL AI systems are both complex and non-intuitive, and these mental models proved to be impressively wrong and surprisingly persistent in many cases. During our pilot tests, we found that setting expectations about the AI system, including general capabilities and expectations for performance, to be critical. Even with these warnings, we found several themes of persistent and incorrect mental models: 

\begin{itemize}
   \item{Multiple participants claimed that the “AI is terrible” and were reticent to use it. To balance our experimental conditions, we intentionally created the AI to perform poorly in about 1/3 of the waves. However, commercial game AI is carefully developed to perform well or as expected in a vast majority of situations. A few key failures of the user study AI and some participants were ready to toss it in the bin and move on.}
   \item{Other participants claimed that the explanations for the AI “made no sense”, i.e., they couldn’t think of a reason that the AI would fail in that way, because a human would never fail in that way. In some cases, participants rejected the correct and consistent explanations claiming this reason, even though the participants understood the explanation. They focused on how a reasonable person might exhibit the failures on the AI.}
    \item{Many participants used metaphors and anthropomorphization to explain the AI behavior, treating it like a human. “It gets overwhelmed.” “It is scared of the big enemies." In other cases, simple behavior (e.g., attacking the nearest enemy) resulted in complex human-like explanations with goals, beliefs, and tactics: “The main thing that I noticed is that it would take off and go for the biggest guy. It would help me split up, but I would have to bail him out or my guys would be spread too thin. I noticed that he was sort of around the larger one, so it looked like he was controlling them individually while I had my guys clustered.”} 
\end{itemize}

Figure \ref{UI} shows the CAMEL explanation interface, which was developed to provide detailed and specific causal explanations for any of the AI actions in recorded episode. To reduce bias and increase participant examination of these explanations, we developed a protocol to introduce participants to each element of the interface, an experimenter-led tutorial to enable them to use the interface, and a walkthrough of each explanation. Participants were asked to read explanations aloud and then reflect upon what they meant, and they did this multiple times before being allowed to explore the many actions and explanations freely. Providing stark contrasting examples of good and bad AI behavior and providing consistent domain-oriented narrative explanations for this behavior helped to correct erroneous participant bias and folk models and improve overall participant mental models of the agent and human-AI teaming on this task.

\subsection{Better Mental Models, Better Teaming}

\begin{figure*}
    \includegraphics[width=500pt]{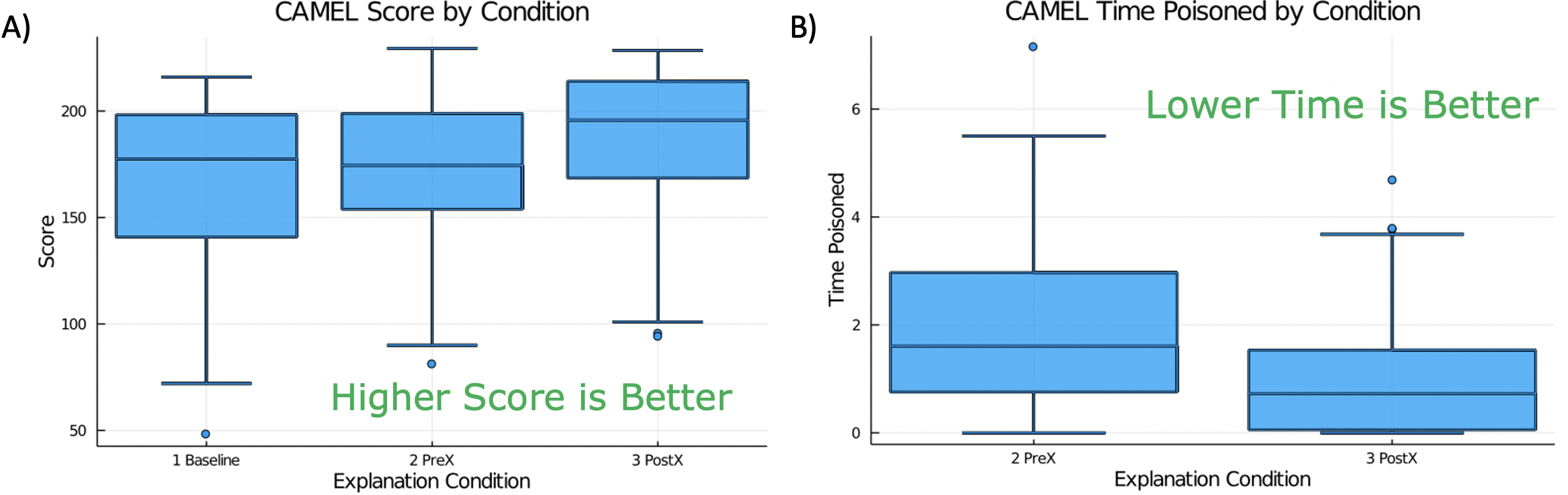}
    \caption{Results from our final human subjects study. Figure A) shows an increases in task performance compared to a baseline. Figure B) shows improved decision maker usage, taking advantage of the AI when it will be high performing, and not using it when it will not.}
    \label{study}
\end{figure*}

A reassuring, if not particularly surprising, finding of the user study was that better mental models led to better teaming. Perhaps more surprising was ways in which they improved performance. Not only were our participants able to better choose when to use and not use the AI, they were also able to accommodate for the AI’s shortcomings when it was activated (see Figure \ref{study}). The user study AIs were “poisoned” to recognize specific situations and perform a failure behavior: stop attacking the enemies and simply move towards them. The poisoning combined position, health, and/or number of units in the environment. For example, one poisoning was “if there is an enemy in the bottom left quadrant and you have an odd number of Marines, fail.” If the participant understood this poisoning, they could use their own unit positioning to draw enemies out of the bottom left quadrant and avoid the poisoned situation. We observed this happening often after participants received CAMEL explanations.

In many domains, people who understand how an AI works are better able to work with it. A simpler, more consistent, and less capable AI may result in better human-machine team performance than a complex, nuanced, and more capable AI. We suggest that AIs trained for human-machine teaming should consider the high costs of complexity and unpredictability, and we observe that one of the key contributions of XAI is to enable more capable AI to effectively team with humans through explanation and understanding.

\section{Conclusions and Future Work}\label{sec5}
Our participation in the DARPA XAI program was illuminating in laying bare the challenges and opportunities in making AI-driven systems more explainable. In this paper we discussed some of our major takeaways from our work with explaining deep RL-driven autonomous systems using causal modeling. The first major takeaway is that good explanations are highly dependent on the user receiving the explanation, and the scenario at hand. The level of technical training by the user, the amount of time they have to process the explanation, the degree to which they need to understand the AI are all of critical importance in supplying effective explanation. The second major takeaway was that one can not always provide meaningful explanations for autonomous agents simply because there fundamentally isn't necessarily anything meaningful to say about their actings because they are simply \emph{reacting} to the particular input sample and have no semantic understanding of their environment. The third major takeaway is that causal models can be fantastic tools for providing explanations, but appropriately using them to provide effective explanations is challenging--two primary challenges we experienced was that: output needs to be appropriately mapped to a vocabulary the user understands; and the plausibility of the explanation needs to be addressed. The final takeaway, and perhaps the most salient, is that users quickly formulate deeply entrenched mental models for AIs, and breaking these incorrect mental models and replacing them with correct ones, requires consistent and compelling explanations. 

Although much was accomplished along the course of the XAI program, from our team and others, it became clear that there are exciting avenues for future work in the area of explainability for AI systems. In our research we focused on \emph{ad hoc} explainability, or augmenting pre-existing systems with explainability. E.g., incorporating human understandable features (rather than low level features such as RGB images), adding terms to the loss functioning penalizing a lack of explainability into the learning procedure to actively promote explainability in the learning procedure, or even modifying the architecture to make it more explainable. In realm of using causal modeling, enhancing the sophistication of the underlying model could add to the complexity of the AIs it could describe. E.g., adding latent feature information (e.g., interim layer activation values from a DNN) could help improve the accuracy. Additionally, improving the method of ranking plausibility is a natural follow up to our research. Finally, a potentially very fruitful area of future research is further improving HMT by improving AI usage via more effective training. Specifically,  adaptive, tailored explanations for particular users would be extremely helpful as AIs take on more and more complicated tasking. The ability to better audit the AI to build better mental models on how it works and when it will fail will be essential. 


\section*{Acknowledgments}
This research was developed with funding from the Defense Advanced Research Projects Agency (DARPA). This material is based on research sponsored by the Air Force Research Lab (AFRL) under agreement number FA8750-17-C-0118. The U.S. Government is authorized to reproduce and distribute reprints for governmental purposes notwithstanding any copyright notation thereon. The views, opinions and/or findings expressed are those of the author and should not be interpreted as representing the official views or policies of the Department of Defense or the U.S. Government.

\bibliography{wileyNJD-AMA}%

\begin{thebibliography}{10}
\providecommand \doibase [0]{http://dx.doi.org/}%

\bibitem{Hirt1974}
Hirt C, Amsden A, Cook J. An arbitrary {L}agrangian-{E}ulerian computing method
  for all flow speeds. {\it J {C}omput {P}hys} 1974\string; 14(3)\string:
  227--253.

\bibitem{Liska2010}
Liska R, Shashkov M, Vachal P, Wendroff B. Optimization-based synchronized
  flux-corrected conservative interpolation (remapping) of mass and momentum
  for arbitrary {L}agrangian-{E}ulerian methods. {\it J {C}omput {P}hys}
  2010\string; 229(5)\string: 1467--1497.

\bibitem{Taylor1937}
Taylor G, Green A. Mechanism of the production of small eddies from large ones.
  {\it P {R}oy {S}oc {L}ond {A} {M}at} 1937\string; 158(895)\string: 499--521.
\newblock \url{https://doi.org/10.1098/rspa.1937.0036},
  \url{http://rspa.royalsocietypublishing.org/content/158/895/499}.

\bibitem{Knupp1999}
Knupp P. Winslow smoothing on two-dimensional unstructured meshes. {\it Eng
  {C}omput} 1999\string; 15\string: 263--268.

\bibitem{Kamm2000}
Kamm J. Evaluation of the {S}edov-von {N}eumann-{T}aylor blast wave solution.
  Tech. Rep. Technical {R}eport LA-UR-00-6055, Los {A}lamos {N}ational
  {L}aboratory;  2000.

\bibitem{Kucharik2003}
Kucharik M, Shashkov M, Wendroff B. An efficient linearity-and-bound-preserving
  remapping method. {\it J {C}omput {P}hys} 2003\string; 188(2)\string:
  462--471.

\bibitem{Blanchard2015}
Blanchard G, Loubere R. High-Order {C}onservative {R}emapping with a posteriori
  {MOOD} stabilization on polygonal meshes. 2015.
\newblock \url{https://hal.archives-ouvertes.fr/hal-01207156}, the {HAL} {O}pen
  {A}rchive, hal-01207156. Accessed January 13, 2016.

\bibitem{Burton2013}
Burton D, Kenamond M, Morgan N, Carney T, Shashkov M. An intersection based
  {ALE} scheme {(xALE)} for cell centered hydrodynamics {(CCH)}. In: Talk at
  {M}ultimat 2013, {I}nternational {C}onference on {N}umerical {M}ethods for
  {M}ulti-{M}aterial {F}luid {F}lows. ; September 2--6, 2013; San {F}rancisco.
\newblock LA-UR-13-26756.2.

\bibitem{Berndt2011}
Berndt M, Breil J, Galera S, Kucharik M, Maire P, Shashkov M. Two-step hybrid
  conservative remapping for multimaterial arbitrary {L}agrangian-{E}ulerian
  methods. {\it J {C}omput {P}hys} 2011\string; 230(17)\string: 6664--6687.

\bibitem{Kucharik2012}
Kucharik M, Shashkov M. One-step hybrid remapping algorithm for multi-material
  arbitrary {L}agrangian-{E}ulerian methods. {\it J {C}omput {P}hys}
  2012\string; 231(7)\string: 2851--2864.

\bibitem{Breil2015}
Breil J, Alcin H, Maire P. A swept intersection-based remapping method for
  axisymmetric {ReALE} computation. {\it Int {J} {N}umer {M}eth {F}l}
  2015\string; 77(11)\string: 694--706.
\newblock Fld.3996.

\bibitem{Barth1997}
Barth T. Numerical methods for gasdynamic systems on unstructured meshes. In:
  Kroner D, Rohde C, Ohlberger M. \kern-2pt, eds. {\it An {I}ntroduction to
  {R}ecent {D}evelopments in {T}heory and {N}umerics for {C}onservation {L}aws,
  {P}roceedings of the {I}nternational {S}chool on {T}heory and {N}umerics for
  {C}onservation {L}aws}Lecture {N}otes in {C}omputational {S}cience and
  {E}ngineering. Berlin: Springer.  1997.
\newblock ISBN 3-540-65081-4.

\bibitem{Lauritzen2011}
Lauritzen P, Erath C, Mittal R. On simplifying `incremental remap'-based
  transport schemes. {\it J {C}omput {P}hys} 2011\string; 230(22)\string:
  7957--7963.

\bibitem{Klima2017}
Klima M, Kucharik M, Shashkov M. Local error analysis and comparison of the
  swept- and intersection-based remapping methods. {\it Commun {C}omput {P}hys}
  2017\string; 21(2)\string: 526--558.

\bibitem{Dukowicz2000}
Dukowicz J, Baumgardner J. Incremental remapping as a transport/advection
  algorithm. {\it J {C}omput {P}hys} 2000\string; 160(1)\string: 318--335.

\bibitem{Kucharik2011}
Kucharik M, Shashkov M. Flux-based approach for conservative remap of
  multi-material quantities in {2D} arbitrary {L}agrangian-{E}ulerian
  simulations. In:  Fo\v{r}t J, F{\"{u}}rst J, Halama J, Herbin R, Hubert F.
  \kern-2pt, eds. {\it Finite {V}olumes for {C}omplex {A}pplications {VI}
  {P}roblems \& {P}erspectives}. 1 of {\it Springer {P}roceedings in
  {M}athematics}. Springer.  2011 (pp. 623--631).

\bibitem{Kucharik2014}
Kucharik M, Shashkov M. Conservative multi-material remap for staggered
  multi-material arbitrary {L}agrangian-{E}ulerian methods. {\it J {C}omput
  {P}hys} 2014\string; 258\string: 268--304.

\bibitem{Loubere2005}
Loubere R, Shashkov M. A subcell remapping method on staggered polygonal grids
  for arbitrary-{L}agrangian-{E}ulerian methods. {\it J {C}omput {P}hys}
  2005\string; 209(1)\string: 105--138.

\bibitem{Caramana1998}
Caramana E, Shashkov M. Elimination of artificial grid distortion and
  hourglass-type motions by means of {L}agrangian subzonal masses and
  pressures. {\it J {C}omput {P}hys} 1998\string; 142(2)\string: 521--561.

\bibitem{Hoch2009}
Hoch P. An arbitrary {L}agrangian-{E}ulerian strategy to solve compressible
  fluid flows. Tech. Rep. Technical {R}eport, CEA;  2009.
\newblock HAL: hal-00366858.
  https://hal.archives-ouvertes.fr/docs/00/36/68/58/PDF/ale2d.pdf. Accessed
  January 13, 2016.

\bibitem{Shashkov1996}
Shashkov M. {\it Conservative {F}inite-{D}ifference {M}ethods on {G}eneral
  {G}rids}.
\newblock Boca Raton, Florida: CRC {P}ress .
\newblock 1996.
\newblock ISBN 0-8493-7375-1.

\bibitem{Benson1992}
Benson D. Computational methods in {L}agrangian and {E}ulerian hydrocodes. {\it
  Comput {M}ethod {A}ppl {M}} 1992\string; 99(2--3)\string: 235--394.

\bibitem{Margolin2003}
Margolin L, Shashkov M. Second-order sign-preserving conservative interpolation
  (remapping) on general grids. {\it J {C}omput {P}hys} 2003\string;
  184(1)\string: 266--298.

\bibitem{Kenamond2013}
Kenamond M, Burton D. Exact intersection remapping of multi-material
  domain-decomposed polygonal meshes. In: Talk at {M}ultimat 2013,
  {I}nternational {C}onference on {N}umerical {M}ethods for {M}ulti-{M}aterial
  {F}luid {F}lows. ; September 2--6, 2013; San {F}rancisco.
\newblock LA-UR-13-26794.

\bibitem{Dukowicz1984}
Dukowicz J. Conservative rezoning (remapping) for general quadrilateral meshes.
  {\it J {C}omput {P}hys} 1984\string; 54(3)\string: 411--424.

\bibitem{Margolin2002}
Margolin L, Shashkov M. Second-order sign-preserving remapping on general
  grids. Tech. Rep. Technical Report LA-UR-02-525, Los {A}lamos {N}ational
  {L}aboratory;  2002.

\bibitem{Mavriplis2003}
Mavriplis D. Revisiting the least-squares procedure for gradient reconstruction
  on unstructured meshes. In: AIAA 2003-3986. 16th {AIAA} {C}omputational
  {F}luid {D}ynamics {C}onference. ; June 23--26, 2003; Orlando, {F}lorida.

\bibitem{Scovazzi2008}
Scovazzi G, Love E, Shashkov M. Multi-scale {L}agrangian shock hydrodynamics on
  {Q1/P0} finite elements: {T}heoretical framework and two-dimensional
  computations. {\it Comput {M}ethod {A}ppl {M}} 2008\string;
  197(9--12)\string: 1056--1079.

\end{thebibliography}


\begin{thebibliography}{10}
\providecommand \doibase [0]{http://dx.doi.org/}%

\bibitem{goodfellow2016deep}
Goodfellow I, Bengio Y, Courville A, Bengio Y. {\it Deep learning}. 1.
\newblock MIT press Cambridge .
\newblock 2016.

\bibitem{mao2016image}
Mao XJ, Shen C, Yang YB. Image restoration using very deep convolutional
  encoder-decoder networks with symmetric skip connections. {\it arXiv preprint
  arXiv:1603.09056} 2016.

\bibitem{zaremba2014recurrent}
Zaremba W, Sutskever I, Vinyals O. Recurrent neural network regularization.
  {\it arXiv preprint arXiv:1409.2329} 2014.

\bibitem{lecun1995convolutional}
LeCun Y, Bengio Y, others . Convolutional networks for images, speech, and time
  series. {\it The handbook of brain theory and neural networks} 1995\string;
  3361(10)\string: 1995.

\bibitem{gunning2017explainable}
Gunning D. Explainable artificial intelligence (xai). {\it Defense Advanced
  Research Projects Agency (DARPA), nd Web} 2017\string; 2(2).

\bibitem{lee2004trust}
Lee JD, See KA. Trust in automation: Designing for appropriate reliance. {\it
  Human factors} 2004\string; 46(1)\string: 50--80.

\bibitem{hoffman2018metrics}
Hoffman RR, Mueller ST, Klein G, Litman J. Metrics for explainable AI:
  Challenges and prospects. {\it arXiv preprint arXiv:1812.04608} 2018.

\bibitem{wity_gen}
Witty S, Lee J, Tosch E. Generalization in Deep Reinforcement Learning. 2018.
\newblock \url{http://www.ctan.org/pkg/algorithm2e}.

\bibitem{vinyals2017starcraft}
Vinyals O, Ewalds T, Bartunov S, et al. Starcraft ii: A new challenge for
  reinforcement learning. {\it arXiv preprint arXiv:1708.04782} 2017.

\bibitem{mnih2013playing}
Mnih V, Kavukcuoglu K, Silver D, et al. Playing atari with deep reinforcement
  learning. {\it arXiv preprint arXiv:1312.5602} 2013.

\bibitem{schrittwieser2020mastering}
Schrittwieser J, Antonoglou I, Hubert T, et al. Mastering atari, go, chess and
  shogi by planning with a learned model. {\it Nature} 2020\string;
  588(7839)\string: 604--609.

\bibitem{cobbe2019quantifying}
Cobbe K, Klimov O, Hesse C, Kim T, Schulman J. Quantifying generalization in
  reinforcement learning. In: PMLR. ; 2019\string: 1282--1289.

\bibitem{atrey2019exploratory}
Atrey A, Clary K, Jensen D. Exploratory not explanatory: Counterfactual
  analysis of saliency maps for deep reinforcement learning. {\it arXiv
  preprint arXiv:1912.05743} 2019.

\bibitem{menzies2001counterfactual}
Menzies P, Beebee H. Counterfactual theories of causation.  2001.

\end{thebibliography}

\clearpage


\end{document}